\documentclass[10pt,twocolumn,letterpaper]{article}

\usepackage{cvpr}
\usepackage{times}
\usepackage{epsfig}
\usepackage{graphicx}
\usepackage{amsmath}
\usepackage{amssymb}
\usepackage{graphics}
\usepackage{graphicx}
\usepackage[]{graphicx} 
\usepackage{epsfig} 
\usepackage{subfigure}
\usepackage{algorithm}
\usepackage{algorithmic}
\usepackage{stmaryrd}
\usepackage[mathscr]{eucal}
\usepackage{lineno}
\usepackage{color}
\usepackage{filecontents}
\usepackage{subfigure}
\usepackage{multirow}
\usepackage{amsmath}
\usepackage{amssymb}
\usepackage{filecontents}
\usepackage{verbatim} 
\usepackage{tabularx}
\usepackage{lineno}
\usepackage{setspace}
\usepackage{multirow}
\usepackage{textcomp,booktabs}

\def\a{{\bf a}}

\def\D{{\bf D}}

\def\I{{\bf I}}

\def\S{{\bf S}}
\def\x{{\bf x}}

\def\u{{\bf u}}

\def\w{{\bf w}}
\def\0{{\bf 0}}
\def\1{{\bf 1}}

\def\RB{{\mathbb R}}

\def\eg{\emph{e.g. }}
\def\ie{\emph{i.e. }}

\def\argmin{\mathop{\rm argmin}}

\def\diag{\mathsf{diag}}

\def\etal{{\em et al.\/}\,}

\usepackage[pagebackref=true,breaklinks=true,letterpaper=true,colorlinks,bookmarks=false]{hyperref}

\cvprfinalcopy 

\def\cvprPaperID{1} 

\ifcvprfinal\fi
\begin{document}

\title{Spatially Aware Dictionary Learning and Coding for  Fossil Pollen Identification}

\author{Shu Kong\textsuperscript{1}, Surangi Punyasena\textsuperscript{2}, Charless Fowlkes\textsuperscript{1}\\
\textsuperscript{1}UC Irvine  \ \ \ \ \ \ \
{\tt\small \{skong2, fowlkes\}@ics.uci.edu}\\
\textsuperscript{2}UIUC  \ \ \ \ \ \ \
{\tt\small punyasena@life.illinois.edu}
}





\maketitle
\begin{abstract}
We propose a robust approach for performing automatic species-level recognition
of fossil pollen grains in microscopy images that exploits both global shape
and local texture characteristics in a patch-based matching methodology. We
introduce a novel criteria for selecting meaningful and discriminative exemplar
patches.  We optimize this function during training using a greedy submodular
function optimization framework that gives a near-optimal solution with bounded
approximation error.  We use these selected exemplars as a dictionary
basis and propose a spatially-aware sparse coding method to match testing
images for identification while maintaining global shape correspondence.
To accelerate the coding process for fast matching, we introduce a relaxed form
that uses spatially-aware soft-thresholding during coding.
Finally, we carry out an experimental study that
demonstrates the effectiveness and efficiency of our exemplar selection and
classification mechanisms, achieving $86.13\%$ accuracy on a difficult
fine-grained species classification task distinguishing three types of
fossil spruce pollen.\footnote{This work was supported by NSF grants DBI-1262547 and IIS-1253538.}

\end{abstract}

\section{Introduction}

\begin{figure*}[t]
\centering
   \includegraphics[width=0.9\linewidth]{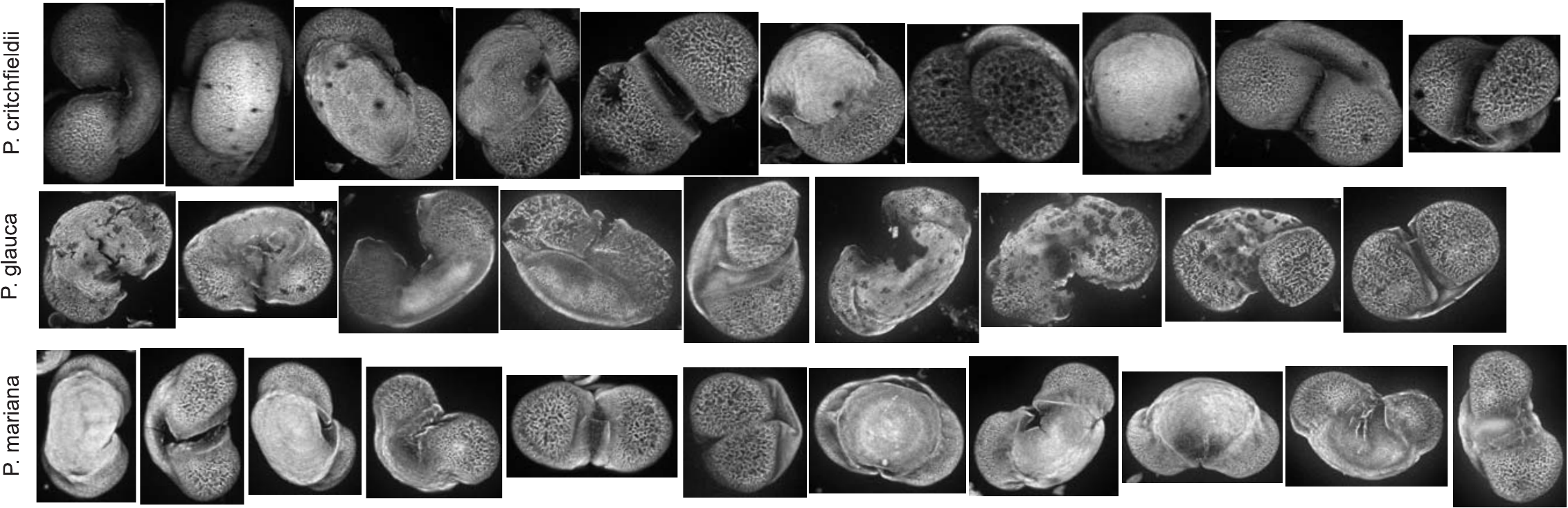}
\vspace{-1mm}
   \caption{Example fossil pollen grains from three species of spruce, imaged
   via confocal fluorescence microscopy.  The fine-grained identification of
   pollen species is not a trivial task and depends on subtle differences in
   the overall pollen grain shape as well as local surface texture.  The
   arbitrary viewpoint, substantial intra-species shape variance and sample
   degradation of the grains poses further difficulties.
   }
\label{fig:examples}
\vspace{-4mm}
\end{figure*}

As one of the most ubiquitous of terrestrial fossils, pollen has an
extraordinarily rich record and has been used to test hypotheses from a broad
cross-section of biological and geological sciences and a diverse array of
disciplines.  Detecting and classifying pollen grains in a collected sample
allows one to estimate the diversity of plant species in a particular area,
carry out paleoecological and paleoclimatological investigations across
hundreds to millions of years, implement the identification of plant speciation
and extinction events, calculate the correlation and biostratigraphic dating of
rock sequences, and conduct studies of long-term anthropogenic impacts on plant
communities and the study of plant-pollinator
relationships~\cite{punyasena2012classifying}.



While high-throughput microscopic imaging allows for ready acquisition of large
numbers of images of modern or fossilized pollen samples, identifying and
counting by eye the number of grains of each species is painstaking work and
requires substantial expertise and training. In this paper, we tackle the
problem of performing automated species-level classification of individual
pollen grains using machine learning techniques based on sparse coding to
capture fine-grained distinctions in surface texture and shape.

A number of previous works have proposed to apply machine learning to pollen
identification~\cite{langford1990computerized,li1999pollen,france2000new,ronneberger2002automated,li2004towards,treloar2004towards,zhang2004towards,chen2006feasibility,dell2009pollen,landsmeer2009detection,holt2011progress}.
However,
all of these methods have largely avoided the difficult problem of species-level classification,
which is significant to the reconstruction of paleoenvironments and discrimination of paleoecologically and paleoclimatically significant taxa~\cite{punyasena2012classifying}.
Recently,
Punyasena \etal proposed two different machine learning-based approaches to identify two pollen species of spruce~\cite{punyasena2012classifying,tcheng2016visual}.
Their approach uses three categories of hand-crafted features,
including intensity distribution, gross shape, and texture features
which are further enriched by varying the parameters for each feature computation.
They show effectiveness of the approach in identifying both modern and fossil pollen grains as a three-way classification problem,
and suggest that the pollen grain size and texture are important variables in pollen species discrimination.
However, they rely on leave-one-out validation to estimate performance and leave open the
question of generalization to held-out test data and other species.



In this paper, we propose a robust framework to automatically identify the
species of fossil pollen grains in microscopy images.  There are several
difficulties that arise, including the arbitrary viewpoint of the pollen grains
imaged (see Fig.~\ref{fig:examples}) and very limited amounts of expert-labeled
training data (relative to many modern computer vision tasks).  To address
these problems, we introduce an exemplar matching strategy for identification
based on local surface patches through several novel technical components.
First, we propose a greedy method for selecting discriminative exemplar patches
based on optimizing a submodular set function.  We show our greedy algorithm is
efficient and gives a near-optimal solution with a $(1-1/e)$-approximation bound.
Second, we use the selected exemplar patches as a codebook dictionary and
propose a spatially aware sparse coding method to match test
image patches for classification.  Finally, to accelerate the matching process for classification,
we introduce a relaxed form of our weighted sparse coding method
for fast matching.  Through experimental study on a dataset of spruce pollen
grains, we demonstrate the efficiency and effectiveness of our patch selection
and classification mechanisms.  Our method achieves $86.13\%$ accuracy on a
three-way classification task which is quite promising given the visual
difficulty of the task and the small training set size.

\section{Discriminative Patch Selection}
In order to allow robust matching of surface texture and local shape features
of pollen grains while maintaining invariance to arbitrary viewpoint (as shown
in Figure~\ref{fig:examples}), we use a patch-based representation of
appearance.  Our first step is to select a small number of exemplar patches
from the training dataset.  The selected patches or exemplars should not only
represent the pollen grains well in the feature space, but also have the
capability to distinguish species-level characteristics by preserving the
spatial structure of the grains.
We use the selected patches as a dictionary basis to match testing images for identification.

To this end, we formulate an objective function that scores a set of candidate
patches selected from the training set based on several criteria including
representational and discriminative power, and balanced sampling across classes
and spatial locations. Selecting a subset of patches that optimizes this
objective reduces to a well studied problem of maximizing a {\em submodular set
function}, which we introduce briefly before describing the specific terms in
our patch selection objective function.

\subsection{Submodular Function Optimization}

Given a finite ground set $\cal V$, a set function ${\cal F}: 2^{\cal V}
\rightarrow \RB$ assigns a value to each possible subset of $\cal V$.
We say ${\cal F}$ is monotonically increasing if ${\cal F}(A) \le {\cal F}(B)$
for all $A \subseteq B$.  A set function ${\cal F}$ is submodular if ${\cal
F}(A\cup a) - {\cal F}(A) \ge {\cal F}(A\cup \{a, b \}) - {\cal F}(A\cup b) $,
for all $A \subseteq {\cal V}$ and $a, b \in {\cal V}/A$.
This is often referred to as {\em diminishing return property}, as the benefit
of adding each additional element to the set decreases as the size of the set
grows.

While maximizing submodular set functions is NP-hard in
general~\cite{cornuejols1983uncapacitated}, a simple heuristic of greedy
forward selection works well in practice and can be shown to have a
$(1-1/e)$-approximation guarantee for monotonic
functions~\cite{cornuejols1983uncapacitated,nemhauser1978analysis}.

%

\subsection{Patch Selection Objective Function}
We generate a large set of candidate patches by sampling randomly and uniformly over spatial
locations across the collection of training images.  The patches could be
represented by pixel values or other features such as
SIFT~\cite{lowe2004distinctive}.  In our experiments, we use activations from
a pretrained CNN as our feature descriptor~\cite{krizhevsky2012imagenet,simonyan2014very}.
We assume the subset of selected patches should be representative of all the
patches in the feature space and yield discriminative compact clusters that are
balanced across classes. In addition patches should be spatially cover most
regions of the pollen grain. We now describe terms that encode each of these
criteria.

\paragraph{Representative in feature space:}
Given a set of $M$ patches which we denote ${\cal V}$, we construct a
$K$-nearest neighbor weighted affinity graph specified by the matrix $\S \in
\RB^{M\times M}$ where $\S_{ij}$ is the similarity (a non-negative value)
between patch $i$ and patch $j$ measured by the Euclidean distance.  Our aim is to select a subset $A \subseteq
{\cal V}$ consisting of patches that are representative in the sense
that every patch in ${\cal V}$ is similar to some patches in the set $A$. We
define the score of a set exemplars $A$ as:
\begin{equation}
{\cal F}_{R}(A)  = \sum_{j \in {\cal V}} \max_{i\in A} {\bf S}_{ij}, \\
\label{eq:term1}
\end{equation}
This function is a monotonically increasing submodular function and can be seen
as a special case of the facility location
problem~\cite{cornuejols1983uncapacitated} where the costs of all the nodes are
the same.

\paragraph{Spatially distributed in input space:}
Similar to the first term ${\cal F}_{R}$ which assures patches are representative
in feature space, we would also like selected patches to be well distributed
spatially in the input training images.  We construct an affinity graph that
stores the proximity of pairs of patches according to their coordinates on the
pollen surface to assure that the selected exemplars to spread over the whole
pollen grain.  We denote this graph similarity matrix by ${\bf L} \in \RB^{M\times
M}$, and formulate it as the following
\begin{equation}
\begin{split}
{\cal F}_{S}(A) = \sum_{j \in {\cal V}} \max_{i\in A} {\bf L}_{ij}
\end{split}
\label{eq:term4}
\end{equation}

\paragraph{Discriminative power:}
Inspired by~\cite{jiang2012submodular}, we adopt a discriminative term to
encourage selection of patches with discriminative power. For a given exemplar
patch $i \in A$, we refer to the $i^{th}$ cluster as the set of all patches
in ${\cal V}$ which are more similar to $i$ than to any other exemplar
$C_i = \{j \in {\cal V} : S_{ij} > S_{kj} \, \forall k \in A / i\}$,
breaking ties arbitrarily.
We measure the discriminative power of such a
clustering based on how pure the clusters are with respect to the category
labels, while favoring a smaller number of clusters, given by:
\begin{equation}
{\cal F}_{D}(A) = \frac{1}{C} \sum_{i\in A} {\max_c N^i_c}  - \vert A \vert,
\label{eq:term2}
\end{equation}
where $N_c^i$ is the number of exemplars from the $c^{th}$ class that are
assigned to the $i^{th}$ cluster,
and $C=\sum_{i \in A} C_i$.
Eq.~\ref{eq:term2} is also a submodular function, and partial proof can be found in~\cite{jiang2012submodular}.

\paragraph{Class balance:}
We further adopt the balancing term introduced in~\cite{kong2014collaborative}
to balance the number of exemplars belonging to different classes:
\begin{equation}
{\cal F}_{B}(A) = \sum_c \log(\vert A_c\vert + 1)
\end{equation}
where $A_c$ is the subset of exemplars in $A$ belonging to class $c$.
The proof can be found in~\cite{kong2014collaborative} that the above term is
monotonically increasing and a submodular function.

\paragraph{Cluster compactness:}
In addition to balancing the size of exemplars of different classes, we
would also like the clusters to be compact so that the total number of clusters
is small and each exemplar represents roughly the same number of patches.  We
utilize the compactness term introduced in~\cite{liu2011entropy} as below:
\begin{equation}
{\cal F}_{C}(A) = -\sum_{i\in A} p(i)\log(p(i)) - \vert A \vert
\end{equation}
where $p(i) = \frac{|C_i|}{|{\cal V}|}$ is the prior probability of a
patch belonging to the $i^{th}$ exemplar cluster.
This is also a submodular function as shown in~\cite{liu2011entropy}.
The above term will also favor a smaller number of clusters.

By combining these terms, our final objective function for selecting patches
is given by:
\begin{equation}
\begin{split}
  {\cal F}(A) \equiv &  \sum\limits_{j=1}^{M} \max_{i\in A} \S_{ij} + \lambda_{S}\sum\limits_{j=1}^{M} \max_{i\in A} {\bf L}_{ij}  \\
  & + \lambda_D \left( \frac{1}{C} \sum_{i\in A} {\max_c N^i_c} - \vert A \vert \right) \\
  & + \lambda_B \sum_c \log(\vert A_c\vert + 1) \\
  & + \lambda_C \left(-\sum_{i\in A} p(i)\log(p(i)) - \vert A \vert \right) 
\end{split}
\label{eq:obj}
\end{equation}
where $\{\lambda_{S},\lambda_{D},\lambda_{B},\lambda_{C}\}$ are hyperparameters
that weigh the relative contribution of each term.  We note that ${\cal
F}(\varnothing)=0$. As each term is a submodular
function, our objective summing up all the five terms is also a submodular function.
Therefore, we can easily use standard
greedy approximation algorithms to approximately maximize the objective
function.

\subsection{Greedy Lazy Forward Selection}
We sketch the naive greedy forward selection algorithm in
Algorithm~\ref{alg:standardGreedy} to maximize our objective function.
It is well known in literature that solving the submodular function by the
greedy algorithm can yield near-optimal solution with a $(1-1/e)$-approximation
bound~\cite{minoux1978accelerated}.
However, while the complexity of this algorithm is linear
in the number of exemplars selected and bounded by $K$,
the computation in each iteration can be very time consuming.
Each update has to recalculate the gains $\Delta$ for all the unselected
patches remaining in ${\cal V}$ which makes direct application of the greedy
method infeasible in practice.

Instead, we utilize the lazy greedy algorithm introduced
in~\cite{minoux1978accelerated} using a max heap structure.
The lazy greedy algorithm, sketched in Algorithm~\ref{alg:lazyGreedy}, maintains an expected
gain for selecting each patch but only recomputes this gain when a patch becomes a candidate
for selection.  This avoids updating many of the gains associated with patches in ${\cal V}$
which are already ``covered'' by an exemplar.
This greedy algorithm with lazy updates is analyzed in~\cite{minoux1978accelerated}
and provides a good approximation to the optimal solution of the NP-hard
optimization problem.
In our experiments,
the lazy greedy Algorithm~\ref{alg:lazyGreedy} yields good solutions and is
hundreds of times faster than the naive greedy
Algorithm~\ref{alg:standardGreedy}.
Specifically, the run time is less than ten minutes to select $K=600$ exemplars
from a pool of $10,000$ candidates on a single CPU.

\renewcommand{\algorithmicrequire}{\textbf{Input:}}
\renewcommand{\algorithmicensure}{\textbf{Output:}}
\begin{algorithm}[t]
\caption{Greedy Selection Algorithm}
\begin{algorithmic}
\small
\REQUIRE ${\cal V}, {\cal F}, K$
\ENSURE a subset $A$ with $\vert A \vert \leq K$
\STATE initialize $A=\varnothing$, $k=0$
\WHILE{$k \leq K$}
\FORALL {$i \in {\cal V} / A$}
  \STATE compute $\Delta(i) = {\cal F}(A\cup\{i\}) - {\cal F}(A)$
\ENDFOR
\STATE $i^*$ = $\arg\max_{i \in {\cal V} / A} \Delta(i)$
\IF{$\Delta(i^*) < 0$}
\RETURN $A$
\ELSE
\STATE $A = A \cup \{i^*\}$, \ \ \ $k =k+1$
\ENDIF
\ENDWHILE
\RETURN $A$
\end{algorithmic}
\label{alg:standardGreedy}
\end{algorithm}

\begin{algorithm}[t]
\caption{Lazy Greedy Selection Algorithm}
\begin{algorithmic}
\small
\REQUIRE ${\cal V}, {\cal F}, K$
\ENSURE a subset $A$ with $\vert A \vert \leq K$
\STATE  initialize $A=\varnothing$, iteration $k=0$
\STATE for all $i \in {\cal V}$, compute $\Delta(i) = {\cal F}(\{i\})$
\WHILE{$k \leq K$}
\STATE $i^*$ = $\arg\max_{i \in {\cal V} / A} \Delta(i)$
\STATE compute $\Delta(i^*) = {\cal F}(A\cup\{i^*\}) - {\cal F}(A)$
\IF{ $\Delta(i^*) \ge \max_{i \in {\cal V} / A} \Delta(i)$ }
        \IF{$\Delta(i^*) < 0$}
        \RETURN $A$
        \ELSE
          \STATE $A = A \cup \{i^*\}$, \ \ \ $k=k+1$
        \ENDIF
\ENDIF
\ENDWHILE
\end{algorithmic}
\label{alg:lazyGreedy}
\end{algorithm}

\section{Spatially Aware Coding for Fast Matching}
The framework of sparse coding has been exploited for a number of computer
vision tasks~\cite{wright2010sparse}, \eg image
classification~\cite{kong2012dictionary} and face
recognition~\cite{wright2009robust}.  In standard coding-based classification,
the individual patch appearance is represented by an abstract code vector
while the spatial location of the patch in a test image is typically ignored.  However, the
spatial coordinates of a patch can be useful to encode
information about the overall shape of a pollen grain and limit comparisons
of local texture between grains to corresponding locations.  Therefore, we propose to
make use of the coordinates in the sparse coding procedure.

\subsection{Spatially-aware Sparse Coding}

Given a dictionary $\D\in \RB^{p\times m}$, one can compute a sparse
representation $\hat \a \in \RB^m$ of a given input $\x \in \RB^p$ over that
dictionary by solving a sparse reconstruction problem:
\begin{equation}
\a^* = \argmin_{\a}
\Vert \x - \D\a \Vert_2^2 + \lambda \Vert \a \Vert_0.
\label{eq:sparseCoding}
\end{equation}
The $\ell_0$ norm $\Vert \cdot\Vert_0$ counts the number of non-zeros in a vector.
We follow the standard approach of replacing this by an $\ell_1$ norm $\Vert
\cdot\Vert_1$ which yields a convex relaxation~\cite{aharon2006img}.

When learning a sparse coding model, it is common practice to learn a
dictionary that is adapted to the dataset by minimizing the reconstruction
error or other discriminative performance measures with respect to the
dictionary elements~\cite{aharon2006img,wright2010sparse}.
In our setup, we use the selected
set of discriminative patch exemplars, as described in the previous section, directly as dictionary elements for
sparse coding-based classification.  One can thus view the selection process as
a discriminative dictionary learning method (see,
e.g.~\cite{kong2012dictionary,ramirez2010classification,mairal2009supervised}).

In order to make use of patch coordinates, we modify the standard sparse coding
objective by including a weight $w_i$ associated with each dictionary element
which encourages codes that are spatially coherent with respect to the training
data.  This weighting can be incorporated into the $\ell_1$ sparsity term
\begin{equation}
\a^*= \argmin_{\a}\Vert \x - \D\a \Vert_2^2 + \lambda_1 \Vert \diag(\w) \a \Vert_1,
\label{eq:SC}
\end{equation}
or alternately by an additional weighted $\ell_2$-norm penalizer
\begin{equation}
\a^*= \argmin_{\a}\Vert \x - \D\a \Vert_2^2 + \lambda_2\Vert \diag(\w) \a\Vert_2^2 + \lambda_1 \Vert \a \Vert_1.
\label{eq:coSC}
\end{equation}
The weight vector $\w$ will depend on the spatial location of the patch $\x$ in
the test image. In particular, $\w_i$ depends on the difference in relative
spatial location of the patch $\x$ and the location of the dictionary atom
(exemplar patch) $i$ in the training image.  Dictionary atoms that were
selected from a very different part of the pollen grain than the patch being
coded are thus more heavily penalized for taking part in the reconstruction.

\begin{figure}[t]
\centering
   \includegraphics[width=0.99\linewidth]{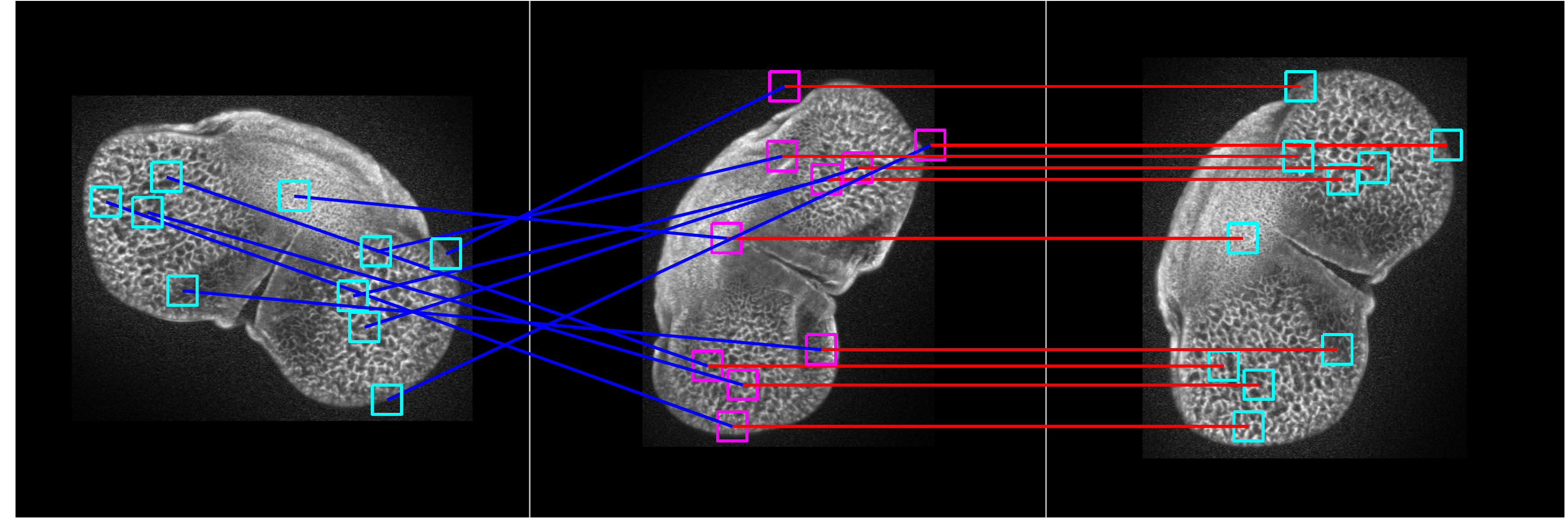}
\vspace{-1mm}
   \caption{The success of our patch-based matching methodology requires that
   the two images are reasonably well aligned w.r.t viewpoint.  We perform
   alignment to remove in-plane rotation and use $k$-medoids clustering to
   group training examples into canonical viewpoints.  After alignment, the
   comparison of patches at corresponding spatial locations between a training
   (center) and test image (right) provides much stronger discriminative
   information than with an unaligned image (left). We exploit this alignment
   by utilizing a spatially adaptive sparse coding scheme we term SACO.}
\label{fig:patchMatchDemo}
\vspace{-2mm}
\end{figure}

\subsection{Fast Spatially-aware Coding}
The spatially-aware sparse coding described above works quite well for
performing classification.  However, as the number of patches sampled in a test
image increases, which will be a case if we desire better classification
performance, the sparse coding process becomes computationally intensive.  To
address this problem, we propose a fast (relaxed) version of spatially aware
sparse coding, which we term SACO for short.

To motivate our approach, suppose we have an under-complete dictionary\footnote{In our experiments this is indeed the case since the patch feature
dimension is larger than the number of exemplar patches} $\D \in \RB^{p\times
m}$, $p\ge m$.  Without the sparsity regularization, the reconstruction problem
$\argmin_{\a}\Vert \x - \D\a \Vert_2^2$ has a simple least-squares
solution given by:
\begin{equation}
\a^* = {\bf \Omega} \x, \text{where } {\bf \Omega}\equiv (\D^T \D) ^{-1}\D^T.
\end{equation}
We thus consider an alternate cost function that seeks a sparse approximation
to the (dense) least-squares code:
\begin{equation}
\a^*= \argmin_{\a}\Vert {\bf \Omega}\x - \a \Vert_2^2 + \lambda_1 \Vert \a \Vert_1
\end{equation}

For orthonormal dictionaries, e.g. as used in wavelet
denoising~\cite{donoho1994ideal,simoncelli1996noise}, ${\bf \Omega} = \D^{-1}$ and
this problem is equivalent to the sparse reconstruction problem.  In general,
it provides an upper-bound on sparse reconstruction since
\begin{equation}
\Vert {\bf \Omega}(\x - \D \a) \Vert_2 \geq \sigma({\bf \Omega}) \Vert \x - \D \a \Vert_2
\end{equation}
where $\sigma({\bf \Omega})$ is a constant that depends on the dimension
and smallest singular value of $\D$.

The primary appeal of this relaxed formulation is that we can easily obtain the
optimal solution by applying a simple soft-thresholding or ``shrinkage''
function independently to each element of the least squares solution:
\begin{equation}
a_i^* = \text{sgn}( u_i) \cdot  \max(0, \vert u_i\vert -\lambda_1), \text{where } \u={\bf \Omega}\x.
\end{equation}

\paragraph{Spatial weighting}
In our problem, suppose we have an under-complete dictionary $\D \in
\RB^{p\times m}$ consisting of the selected patches and precompute
corresponding pseudo-inverse ${\bf \Omega}$.  We then solve the
spatially-weighted variant corresponding to Eq.~\ref{eq:SC} by
\begin{equation}
\a^*= \argmin_{\a}\Vert {\bf \Omega}\x - \a \Vert_2^2 + \lambda_1 \Vert \diag(\w) \a \Vert_1,
\end{equation}
The solution is then given by $\a^*$ whose $i^{th}$ element is the following:
\begin{equation}
a_i^* =  \text{sgn}( u_i) \cdot  \max(0, \vert u_i\vert -\lambda_1w_i), \text{where } \u={\bf \Omega}\x
\end{equation}
We term this scheme SACO-I.

Alternatively, for the counterpart of the $\ell_2$ weighting used in
Eq.\ref{eq:coSC} we have
\begin{equation}
\begin{split}
{\bf \Omega} \equiv & (\D^T \D + \lambda_2\diag(\w)^2) ^{-1}\D^T \\
\u = & {\bf \Omega}\x \\
a_i^* = & \text{sgn}( u_i ) \cdot \max(0, \vert u_i\vert -\lambda_1)\\
\a^* = &[a_1^*, \dots, a_i^*, \dots, a_m^*]^T.
\end{split}
\label{eq:variant2}
\end{equation}
We term this scheme SACO-II.

Both versions of spatial structure aware shrink coding (SACO) enable us to do
the coding in a feed-forward way without iterative optimization required by
sparse reconstruction. This makes the classification process significantly
more efficient than full reconstructive sparse coding.  We find that in
practice, using a non-overcomplete dictionary is not a limitation and that the
SACO approximation leads to very good classification performance in our
experiments.  SACO-I has additional computational advantage over SACO-II
as it avoids inverting a different matrix at each patch location. This
makes it feasible to perform coding densely over the whole image by performing
correlation over the whole image feature map with each element of ${\bf \Omega}$
followed by application of a spatially varying shrinkage function.
Beyond SACO, we utilize global pooling and linear SVM for classification,
as detailed in the next section.



\section{Implementation Details}

\subsection{$k$-medoids Clustering for Viewpoint Alignment}
As demonstrated by Figure~\ref{fig:patchMatchDemo},
the success of our spatially-aware patch-based matching methodology lies in that the images are well aligned w.r.t viewpoint.
To align the images,
we perform unsupervised pre-processing of both training and test images in order to
automatically improve alignment.

\begin{figure}[t]
\centering
   \includegraphics[width=0.99\linewidth]{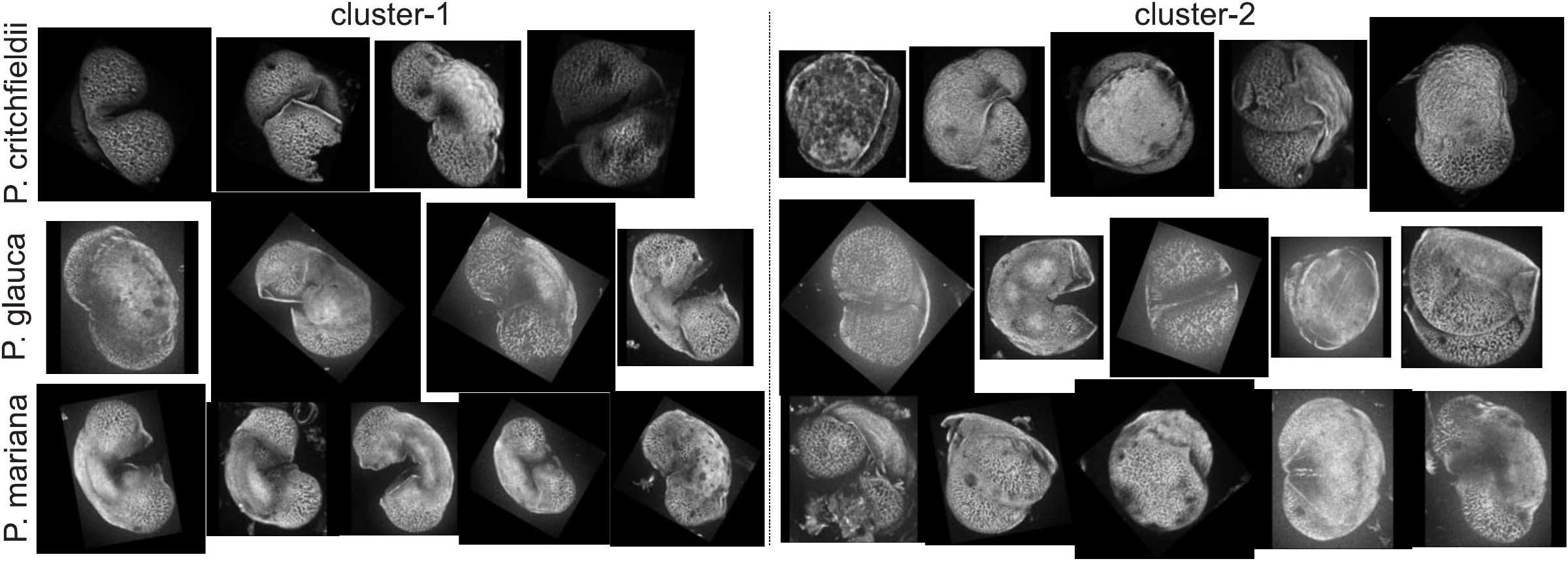}
\vspace{-1mm}
   \caption{Rotated images according to two canonical viewpoints determined by $k$-medoids clustering.
   }
\label{fig:kmedoidDemo}
\vspace{-5mm}
\end{figure}

We use the all the training images (ignoring the species labels) to build an
affinity graph, where the similarity of image $\I_A$ and $\I_B$ is measured by
\begin{equation}
similarity(\I_A,\I_B) =  \frac{1}{\min_\theta \Vert \I_A - R_\theta(\I_B)\Vert},
\end{equation}
where $R_\theta(\I_B)$ is an operator that rotates image $\I_B$ by $\theta$ degrees.
We resize all images and rotated intermediates to $40\times40$ pixel resolution,
and calculate the distance between two images as the sum of squared pixel-wise differences.
We use the resulting similarity graph to perform $k$-medoids clustering.
Empirically, we find that once in-plane rotation is removed, clustering the
images into two canonical viewpoints is enough to achieve good performance.
Figure~\ref{fig:kmedoidDemo} shows two viewpoint clusters with examples from
all three species. Equatorial views of pollen grains appear in the first
cluster while top-down views are assigned to the second cluster.

\subsection{Classification Pipeline}
Using the dictionary $\D \in \RB^{p\times m}$ constructed by the selected
exemplar patches from training images, we perform spatially-aware aware coding
(SACO) for patches of the test image, resulting in $m$-dimensional sparse codes
for each patch.  For each test image we use $50$ patches sampled at random.  We
pool the $m$-dimensional code vectors over the entire set of test patches using
average pooling to produce a final $m$-dimensional feature vector which is fed
into a linear SVM classifier to predict the species.

Rather than using raw image pixel values, we use a feature vector extracted via
the pretrained VGG19 model of~\cite{simonyan2014very}.  We found that using the
features at layer-$conv4\_3$ of VGG19 performed best.  We also analyzed the
performance of SIFT and features at other layers of VGG19 in our experiments
(Section~\ref{sec:exp}).  The receptive field at this layer spans a patch of
$52\times 52$ pixels in the original image.  Figure~\ref{fig:patchMatchDemo}
shows a visualization of these selected patches relative to the scale of the
pollen grain and shows qualitatively that patches capture meaningful local
textures.

\section{Experiments}
\label{sec:exp}
In this section, we introduce our dataset, show the effectiveness of the
proposed exemplar selection method on synthetic data, study different features
used for classification and several hyper-parameters in our pipeline, and
report the classification performance of our models and comparisons to several
strong baselines.

\subsection{Dataset}

We test our method on samples of fossil pollen from three species of spruce,
\emph{Picea critchfieldii}, \emph{Picea glauca}, and \emph{Piciea mariana}.
Samples were chemically extracted from lake sediments as detailed in~\cite{punyasena2012classifying,mander2014identifying}
and imaged using a Zeiss Apotome fluorescence microscope (a form of
structured illumination)~\cite{weigel2009resolution} to produce
high-resolution, three-dimensional image stacks.
Imaging was carried out by multiple operators, with no single person responsible for a single species.
The full shape of the grain was captured by multiple $z$-focal planes at intervals of half the Nyquist frequency~\cite{punyasena2012classifying}.
Grains were cropped manually, using a bounding box that reached from the
maximum width of the grain in the $x$ axis and the maximum length of the grain
in the $y$-axis. The $z$-stack is limited to the uppermost and lowermost
in-focus planes of the grain.
Details of the imaging procedure can be found in [anon].
For each grain, we use maximum intensity projection over the top half of the
grain to produce a single in focus 2D image.  Some examples are show in
Figure~\ref{fig:examples}.

Experts provided a nominal species label for each grain along with a confidence
score. We note that unlike some other image classification tasks, there is no
``ground-truth'' for species identification. However, fossil pollen grains were
taken from strata containing other macro-fossil evidence (e.g., leaves) of
these species and we restricted our analysis to samples with high-confidence
labels.  We randomly split the dataset into training and testing (validation)
sets (statistics are listed in Table~\ref{tab:Dataset}).  The dataset will be
released to public in the near future.

\begin{table}[t]
\footnotesize
\centering
\caption{Statistics of our fossil pollen grain dataset.}
\begin{tabular}{|l|c|c|c|c|}
\hline
                        &	$\#$train &		$\#$test & $\#$total          \\
\hline
\emph{P. critchfieldii} &65            &43  & 108       \\
\emph{P. glauca}        &65            &355 & 420       \\
\emph{P. mariana}       &65            &287 & 352       \\
\hline
Summary                 &195           &685 & 880       \\
\hline
\end{tabular}
\label{tab:Dataset}
\vspace{-5mm}
\end{table}


\subsection{Exemplar Selection on Synthetic Data}
\label{ssec:synthesisExemplar}
\begin{figure}[t]
\centering
   \includegraphics[width=0.99\linewidth,trim={0 0cm 0 0},clip]{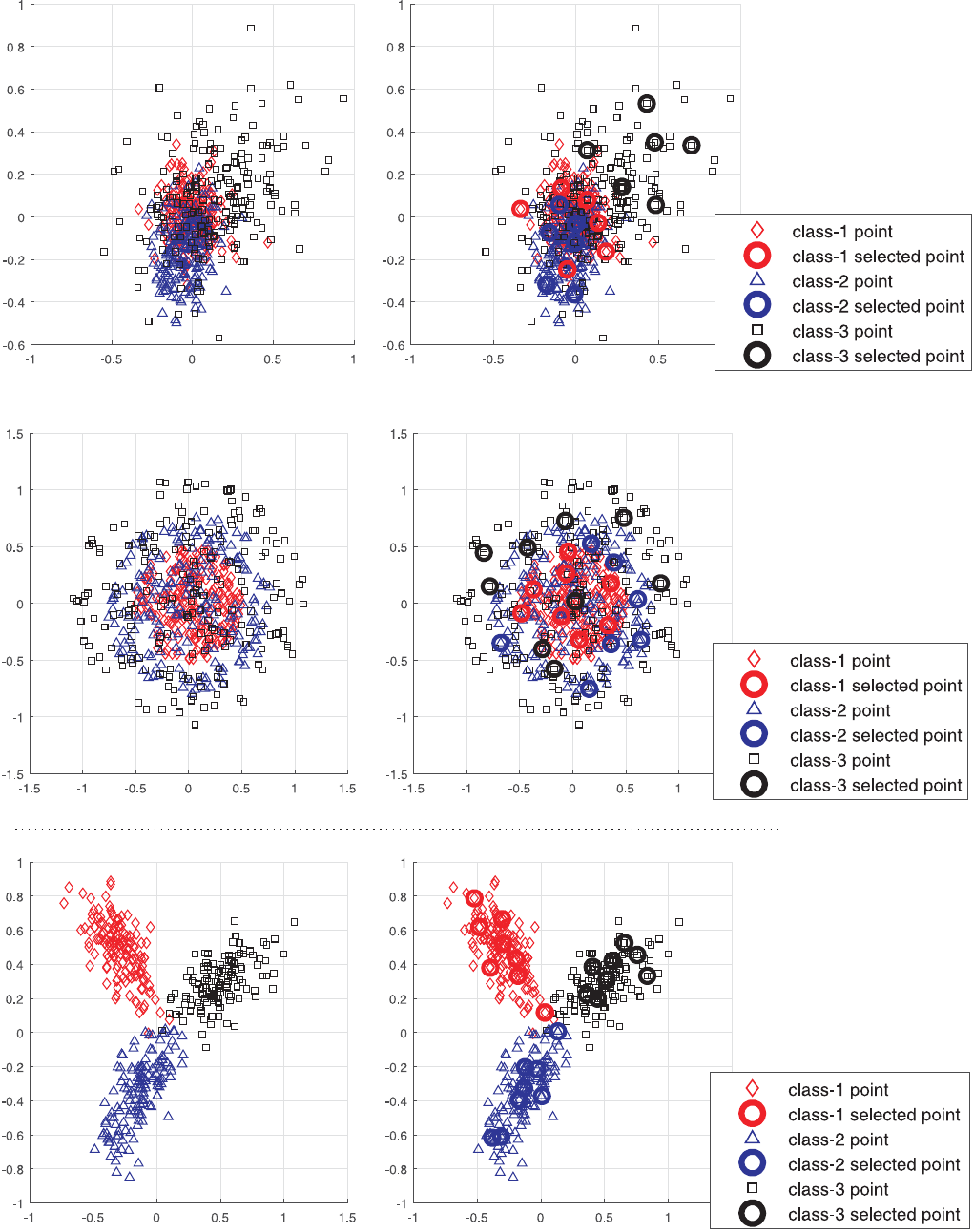}
\vspace{-1mm}
   \caption{Qualitative demonstration of the effectiveness of the proposed
   method in exemplar selection on synthetic data (best seen in color and zoom-in)}
\vspace{-2mm}
\label{fig:synthesisDemo}
\end{figure}

We first verified the effectiveness of the proposed exemplar selection method
using synthetic toy data for which we can easily visualize the results
qualitatively.  In this setting we merge term  ${\cal F}_R$ and  ${\cal F}_S$ in the objective
function (Eq.~\ref{eq:obj}) to simplify our analysis as the 2D data themselves are
both features and spatial coordinates.
As can be seen in Figure~\ref{fig:synthesisDemo}, the greedy lazy
algorithm selects representative exemplars that cover the data points from
each class while maintaining discriminative power by sampling near class
boundaries but avoid non-discriminative areas of feature space with high
inter-class overlap.


\subsection{Choice of Feature Representation}
\begin{figure}[t]
\centering
   \includegraphics[width=0.70\linewidth]{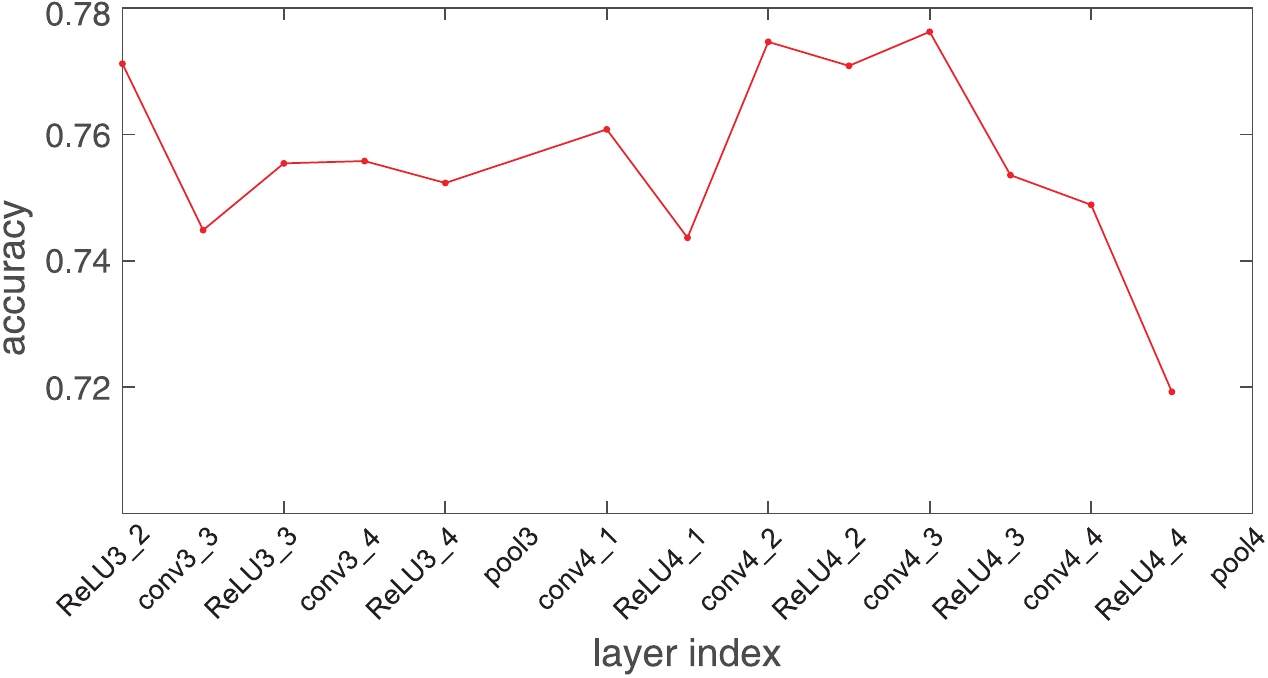}
\vspace{-0.5mm}
   \caption{Classification accuracy vs. layer index in VGG19 model. We use
   features extracted from $conv4\_3$ in the remainder of our experiments.}
\label{fig:acc_vs_CNNlayer}
\vspace{-3mm}
\end{figure}

To study the choice of feature representation, we compare the performance of
SIFT descriptors~\cite{lowe2004distinctive} and features extracted at different
layers of VGG19 model~\cite{simonyan2014very}.  Using SACO-I with SIFT feature
yields $54.40\%$ classification accuracy while CNN features perform much
better.  Figure~\ref{fig:acc_vs_CNNlayer} shows performance of features
extracted at different layers of the VGG19 hierarchy with layer $conv4\_3$
achieving a performance at $77.62\%$ classification accuracy. Receptive fields
at this layer span $52\times52$ pixel patches in the original image and are
visualized in in Figure~\ref{fig:patchMatchDemo}.  We use these features in our
remaining classification experiments.

%


%

\subsection{Evaluation of Dictionary Learning}

\begin{figure}[t]
\centering
   \includegraphics[width=0.95\linewidth]{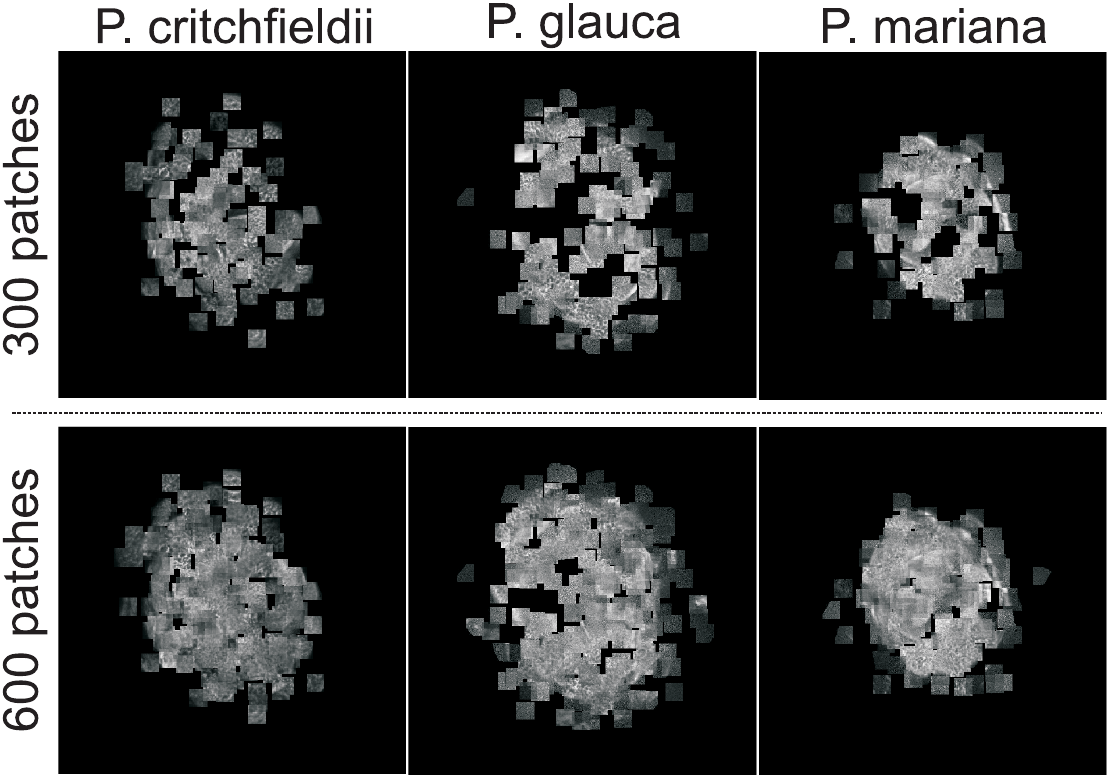}
\vspace{-1mm}
   \caption{To visualize the selected patches, we paste patches selected from
   the same species on a black background to the places according to their
   coordinates.  The two panels show the 300 and 600 patches respectively.
   Our dictionary selection approach favors patches that cover the pollen grain
   spatially, focusing on more discriminative regions when the number of
   dictionary elements is limited (300 patches) but eventually covering the
   whole grain at larger dictionary sizes (600 patches).
   }
\label{fig:visualizeExemplars}
\vspace{-3mm}
\end{figure}

\begin{table}[t]
\small
\centering
\begin{tabular}{|l|c|c|c|c|}
\hline
dictionary size                        &	300      &		512 & 600          \\
\hline
Random Selection & $77.66$     &$76.49$ & $77.23$       \\
\hline
Discriminative Selection & $81.75$     &$81.60$ & $82.34$       \\
\hline
\end{tabular}
\caption{Classification accuracy ($\%$) for different sized dictionaries
constructed by our discriminative exemplar selection algorithm. Our method
consistently outperforms a baseline that selects patches at random from
the training set.}
\label{tab:acc_vs_dictSize}
\vspace{-3mm}
\end{table}

In addition to the synthetic tests in Section~\ref{ssec:synthesisExemplar}, we
verify the effectiveness of our exemplar selection method in the pollen
identification task by comparing the classification performance of dictionaries
consisting of randomly sampled patches. We also report the performance as a
function of varying dictionary size.

We use SACO-I for this experiment, and vary the dictionary size by (randomly)
selecting 300, 512 and 600 patches.  The results are listed in
Table~\ref{tab:acc_vs_dictSize}.  First, it is clear that a dictionary built
from our selected exemplars performs much better than the counterpart
consisting of randomly sampled patches.  Second, a smaller dictionary of 300
atoms is sufficient for our classification task.  However, it appears larger
dictionaries do not harm performance.  We expect that some hyper-parameters
will have an important impact on the performance for larger dictionaries, in
particular regularization of the SVM.  We study the effect of hyper-parameters
in Section~\ref{ssec:paramStudy}.  We use a 300-basis dictionary for the rest
of our experiments.

To visualize the selected patches in the dictionary, we paste them on a black
panel according to their coordinates.  Figure~\ref{fig:visualizeExemplars}
shows the patches of the three species.  We can see that these patches not only
capture local texture information, but also convey a global shape and average
size of the three species.

%
%
%

\subsection{Comparison of SACO Methods and Baselines}
\label{ssec:twoVariants}

\begin{table}[t]
\small
  \centering
  \begin{tabular}{ | c| c| c | c|  c|  c| }
     \hline
        SRC & VGG19+SVM & FV+SVM & SACO-I & SACO-II \\
     \hline
     62.04 & 65.11 & 61.46 & 83.21 & 86.13 \\
     \hline
   \end{tabular}
\caption{Performance of baselines and our SACO methods measured by classification accuracy ($\%$).}
\label{tab:twoVariants}
\vspace{-3mm}
\end{table}


\begin{figure*}[ht!]
\centering
   \includegraphics[width=0.850\linewidth]{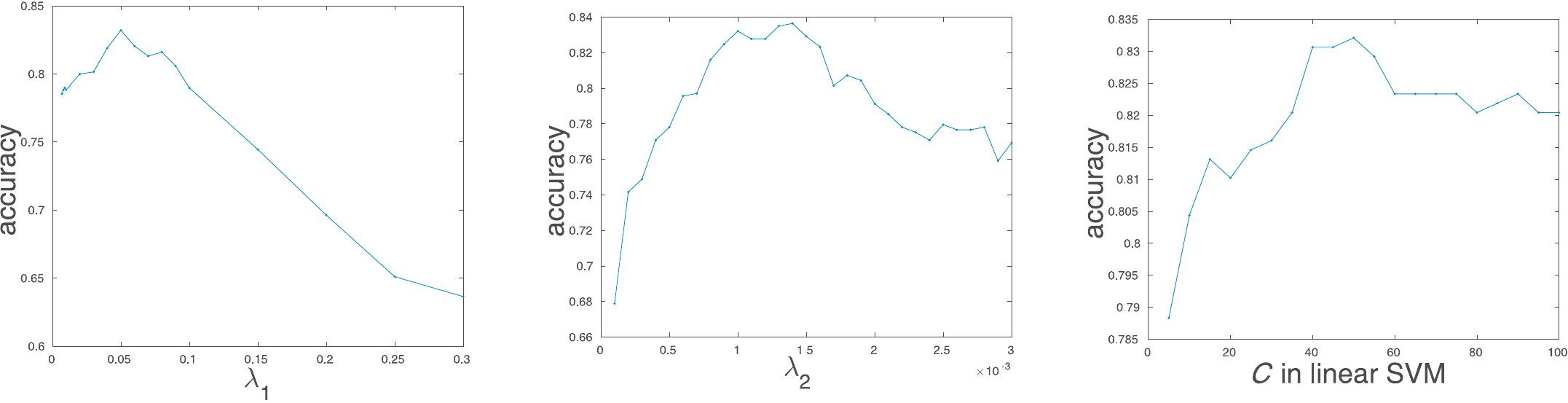}
\vspace{-1mm}
   \caption{The effect of sparisty
   ($\lambda_1$ in Eq.~\ref{eq:variant2}),
   spatial weighting ($\lambda_2$ in Eq.~\ref{eq:variant2}),
   and SVM regularization ($C$) parameter on performance.  }
\label{fig:paramStudy}
\vspace{-3mm}
\end{figure*}

We report the classification performance of our proposed two variants of SACO
along with several baselines in Table~\ref{tab:twoVariants}.  ``SRC'' is the
sparse-representation based classification method of~\cite{wright2009robust}
using the reconstruction error to identify species.  ``VGG19+SVM'' is a
standard CNN-based image classification approach that applies the VGG19 model
to the whole image and performs classification using a linear SVM applied to
features from a high-level layer~\cite{donahue2013decaf}.  ``FV+SVM'' uses
Fisher Vector~\cite{perronnin2010improving} to pool features at a specific
layer of VGG19 to represent the entire image, and applies an SVM
classifier~\cite{cimpoi2015deep}.  For the VGG19 baselines, we tried features
at different layers of VGG19, and report the best result here.

VGG19+SVM provides a strong baseline for shape-based object recognition while
FV+SVM has shown strong performance on texture
classification~\cite{cimpoi2015deep}.
However, neither of these standard methods is competitive with SACO.
It is worth noting that, if fine-tuning the VGG model with softmax loss,
we only obtain $52.41\%$ accuracy. We posit two reasons.
First the original images are of high resolution, so it is easy to overfit the training set.
Second, if we down size the images, the valuable textural characteristics will be eliminated.
SRC is closer to our approach but also performs significantly worse.  When using
random patches without the spatial information as a dictionary in SRC, we only
achieve an accuracy of $57.12\%$.  Adding spatial information and using
selected patches in SRC improves performance to $62.04\%$.  Performing average
pooling over the sparse codes, \ie using our SACO-I method, provides a
substantial improvement in performance, reaching $83.21\%$ classification
accuracy.  The alternative method SACO-II yields even better performance,
$86.13\%$.  This shows the importance of the spatial information of patches in
our task and the need to fuse both shape and texture cues.


\subsection{Parameter sensitivity}
\label{ssec:paramStudy}

There are several important hyper-parameters in our pipeline, including the
sparsity $\lambda_1$, spatial weighting $\lambda_2$ (see
Eq.~\ref{eq:variant2}), and regularization parameter $C$ in the linear SVM.
Figure~\ref{fig:paramStudy} shows accuracy as a function of each of these
hyper-parameters.  The curve showing accuracy as a function of $\lambda_1$
shows that inducing sparsity improves classification performance notably.  The
second curve showing accuracy vs. $\lambda_2$, makes it clear that
incorporating spatially-varying weights on the dictionary elements also
improves the classification performance remarkably.  However, it is necessary
to jointly tune both $\lambda_1$ and $\lambda_2$ for best performance.  Last,
we note the performance is stable w.r.t the parameter $C$ in linear SVM over a
large range.

\subsection{Dense convolutional SACO}
An intriguing aspect of the SACO-I formulation is that it is amenable to a
dense implementation that performs coding at every patch location in the test
image.  This is implemented by first correlating the input image or feature map
with each element of the pseudo-inverse dictionary ${\bf \Omega}$ followed by
soft-thresholding of each response map with a spatially varying threshold and
pooling the result. In theory SACO-II could also be applied densely but demands
significantly more computation since ${\bf \Omega}$ is spatially varying and would
require computing the matrix inverse (Eq.~\ref{eq:variant2}) at every location.

Using a fully convolutional implementation of SACO-I achieved $83.86\%$
classification accuracy.  Although we do not see significant
improvement over the sparse sampling of test patches, we believe better
performance may ultimately be achieved in the dense evaluation by incorporating
automatic segmentation of the pollen grain from background noise and masking of
uninformative damaged areas.  We plan to explore these possibilities in
future work.

\section{Conclusion and Future Work}
We propose a robust framework for pollen grain identification by matching
testing images with a set of discriminative patches selected beforehand from a
training set.  To select the discriminative patches, we introduce a novel
selection approach based on submodular maximization, which is very efficient
and effective  in practice.  To identify pollen grains using the selected
patches as a dictionary, we present two spatially-aware sparse coding methods.
We further accelerate these two methods using a relaxed formulation that
can be computed in an efficient non-iterative manner.

As our experiments show, this spatially aware exemplar-based coding approach
significantly outperforms strong baselines built on state-of-the-art CNN
features.  We leave open as future work the question of how such a matching
mechanism could be fully embedded in a neural network architecture,
how to exploit confidence scores provided with expert labels, and extending the
approach to perform cross-domain matching of fossil and modern pollen samples.


{\small
\bibliographystyle{ieee}
\bibliography{egbib}
}

\end{document}